# Automatic lexical semantic classification of nouns

**Núria Bel, Lauren Romeo, Muntsa Padró**

Universitat Pompeu Fabra
Roc Boronat, 138, Barcelona, Spain
E-mail: {nuria.bel,lauren.romeo,muntsa.padro}@upf.edu

**Abstract**

The work we present here addresses cue-based noun classification in English and Spanish. Its main objective is to automatically acquire lexical semantic information by classifying nouns into previously known noun lexical classes. This is achieved by using particular aspects of linguistic contexts as cues that identify a specific lexical class. Here we concentrate on the task of identifying such cues and the theoretical background that allows for an assessment of the complexity of the task. The results show that, despite of the a-priori complexity of the task, cue-based classification is a useful tool in the automatic acquisition of lexical semantic classes.

## 1. Introduction

Nominal lexical semantic classes gather together properties that appear to be linguistically significant for a number of linguistic phenomena. Determiner selection, selectional restrictions or noun collocation have been described in terms of such groupings of properties. Besides, these classes are often used to generalize over particular senses of different words. For instance, Miller et al. (1990) used a number of lexical semantic classes as features that ordered the nominal meaning hierarchy in WordNet. Applications that use nouns annotated with lexical semantic classes include: machine translation, discrimination of referents in tasks such as event detection and tracking (Fillmore et al., 2006), question answering (Lee et al., 2001), entity typing in named entity recognition (Ciaramita & Altun, 2005; Fu, 2009), automatic building and extending of ontologies (Buitelaar et al., 2005), textual inference (de Marneffe et al., 2009), etc. Furthermore, nominal lexical semantic classes have also recently proved to be useful information for grammar induction (Agirre et al., 2011), where problems come from the need of generalizing over a high dimensional space.

Lexical semantic noun tagging in large lexica is still mostly done by hand, and the high cost of this exercise hinders the production of rich lexica for different languages. In addition, domain tuning of lexica is considered too expensive, and the use of an inadequate lexicon is one of the causes of poor performance of many applications. Thus, current research on automatic production of class-annotated lexica is expected to have a high impact on the performance of most NLP applications. Most critically, it will bring significant improvements in their coverage over different languages and domains. The task appears to be complex, but any reduction in the amount of human work required for the production of these resources can contribute to improve the current situation. This is the ultimate goal of the work we present here.

The work we present addresses cue-based noun classification. Its main objective is to automatically acquire lexical semantic information by classifying nouns into previously known lexical classes. This is achieved by using particular aspects of linguistic contexts where the nouns occur as cues that represent distributional characteristics of a specific lexical class and which also support the building of specialized classifiers. In this research, we have approached lexical semantic information by assuming it to form classes such as EVENT, HUMAN, CONCRETE, SEMIOTIC, LOCATION and MATTER, and we have focused on the task of identifying these class-specific cues: morphological, syntactic and lexical co-occurrence features that provide indicative hints of a particular class.

The limitations of this approach are related to the well-known problem of data sparseness, referring in our case to both the low frequency of most of the words to classify and of the particular cues needed to classify them.

Determining a useful feature set is the most important task for cue-based lexical classification. On the one hand, cues have to be discriminative of the class while on the other hand they must also be frequent enough to be taken into account first by the learner and later by the classifier. In what follows, we present an overview of the theoretical assumptions we have made for defining a methodology to identify such useful features that are successfully used by a classifier.

This paper is organized as follows: in section 2, we discuss cue-based word classification and related work in more detail. In section 3, we describe the theoretical background and methodology used. In section 4, we describe our approach to identify cues. In section 5, we discuss the experiments carried out. In section 6, we present our results and evaluation. Finally, section 7, contains our conclusions and some suggestions for future research.

## 2. Cue-based word classification

Cue-based word classification is based on Harris (1954)

distributional hypothesis formulated for our purpose as follows: words that occur in the same contexts can be said to belong to the same class (Baroni & Lenci, 2010 for an overview on different research lines based on the distributional hypothesis). In addition, Merlo and Stevenson (2001) demonstrated the usefulness of the "markedness" theory that was proposed by the structuralist Prague School linguist Trubetzkoy (1931), which relates the existence of linguistic classes with surface marks. The existence of a class should be "marked" in the sense that there must be formal evidence that characterizes the linguistically motivated class in terms of morphological, syntactic and lexical co-occurrence behaviour (Bybee, 2010).

In this framework, automatic classification is approached by training a learner with information about word occurrences in a selected number of contexts: identified marks or linguistic cues. A learner is supplied with pre-classified words represented by numerical information about matched and not matched cues. The information gathered from the entire set of occurrences of a word in a corpus (that is, the information of all tokens) is taken as evidence to assign the class membership of the type, because the word is observed in a number of particular contexts and because it is not observed in others.

Different supervised Machine Learning (ML) techniques and cue selection methods have been applied to cue-based lexical acquisition. Merlo and Stevenson (2001) used Decision Trees (DT) and selected ad hoc linguistic cues to classify English verbs into three lexical classes: unaccusative, unergative and object-drop. More general cues, such as the part of speech tags of neighbouring words, were proposed by Baldwin (2005), which used a memory based classifier (Daelemans et al., 2003) and Joanis et al. (2007), which used Support Vector Machines (SVM) and the frequency of filled syntactic positions or slots, tense and voice features, etc., as general cues to classify English verbs into Levin classes (Levin, 1993). Joanis et al. (2007) carried out a number of different experiments to identify which types of features were more informative for verb classification. They concluded that syntactic information about predicate complements, as well as prepositions, were most informative.

For noun classification in particular, Hindle (1990) used a similarity metric derived from the shared distribution of subjects, verbs and objects as observed in a corpus. Lin (1998) empirically demonstrated that prepositions and modifiers are quality cues for lexical semantic class identification also based on a similarity metric. Light (1996) successfully used the information from derivational affixes to semantically classify nouns. Baldwin and Bond (2003) used linguistic cues such as co-occurrence with particular determiners, number, etc., to learn the countability of English nouns and trained a memory-based classification system based on the k-nearest neighbour algorithm (TiMBL, Daelemans et al., 2003) with results around 90% of accuracy. Bel et al. (2007) proposed a number of cues for classifying nouns into different classes according to an HPSG-based lexical typology using DT. Bel et al. (2010) dealt with Spanish and English non-deverbal event nouns. They used DT and a few linguistically motivated cues to achieve results with around 80% of accuracy.

From the previous research mentioned, it seems clear that it is possible to rely on a number of contexts to contribute information for classification. Our hypothesis is that it can also be a method for lexical semantic classification.

## 3. Theoretical background and Methodology

Our approach to lexical semantic classification is based on the assumption that lexical classes, including lexical semantic classes, exist. Following previous research, we approach lexical classification assuming, in addition to the classical distributional hypothesis (Harris, 1954), that lexical semantic classes are emergent properties of a number of words that recurrently co-occur in a number of particular contexts, as Bybee and Hopper (2001) and Bybee (2010) propose. Our method also assumes a representational uniformity (Bybee, 2010; Jurafsky, 1996), which is, that all levels of linguistic representation are involved in the assessment of the class. In our work, we resorted to the broadest notion of context, i.e. different suffixes, occurrence in particular grammatical functions and lexical co-occurrence with particular predicates and adjuncts.

In the framework of usage based grammar theories (Goldberg, 2006) and supported by psycholinguistic and cross-lingual evidence, a lexical class is a generalization that comes about when there is a systematic co-distribution for a number of words and a number of contexts in the broad sense just mentioned. Thus, different contexts where a number of words tend to occur become overt linguistic cues of a particular semantic property that a set of words has in common and, therefore, upon which members of that class can be recognized. In other words, nouns belonging to a class will tend to show up in a number of particular contexts. For example, the nouns that are members of the class EVENT will tend to co-occur with prepositions that refer to duration, i.e. *during*, and also with verbs whose meaning refers to events, i.e. *to last*. Construction based grammar hypotheses allow us to predict that there are a set of word occurrences, not in one or another discriminating context, but it is a number of them what constitutes a class mark. The structuralist notion of markedness (Jakobson, 1971 and Bybee, 2010, for a revision) allows for principled predictions about the probability of observing these contexts if understood as class marks (Merlo and Stevenson, 2001).

First, the markedness notion is based on the existence of binary oppositions. Accordingly, we assumed that each lexical semantic class is an independent opposition where marks signal having a semantic feature (Battistella, 1990). Thus, the markedness notion would allow us to predict that marked elements, i.e. members of the class, will appear in marking contexts, but also in non-marking contexts. The absence of marks can be interpreted either as an instance of a non-member, or a situation where the distinction is irrelevant (Jakobson, 1971).

Second, according to Bybee (2010) unmarked elements are more frequent than marked elements. Thus, it is important that the learner correctly assesses the discriminating nature of low frequent phenomena, that is, of the marking class contexts versus the unmarked contexts. Given that supervised machine-learning methods are, in general, insensitive to low frequency (Bel, 2010), which they often assimilate to noise, the induction of relevant cues for classification has problems because low frequency evidence is disregarded. Smoothing methods have proved to raise accuracy; yet, with low frequent words the problem remains because if evidence, i.e. the number of occurrences, is very low, it is not considered as a positive cue for the class. Classification results depend on the identification of these contexts, but also on their availability with the word to classify, given a particular corpus. Therefore, we assumed that only frequently occurring contexts are going to be efficient for our task; thus, we looked for frequent predicates, prepositions, etc. as possible indicators for a particular class. Note, that we do not count on having unique, exclusive hints for a class, but a number of them that, when correlated, can identify the members of the class.

In addition, co-occurrence of a noun with a predicate as its complement is, in principle, subject to selectional restrictions, i.e. semantic requisites that regulate semantic compositional analysis. However, Pustejovsky and Ježek (2008) worked out the limits of class distribution for meaning distinctions. Contextual phenomena like coercion can alter the distributional behaviour of words allowing them to show up in contexts that do not correspond uniquely to their "inherent" semantic properties. Within the framework of the Generative Lexicon, Pustejovsky and Ježek (2008) resort to the idea that some predicates not only select, but also even impose these restrictions to the filler of the complement. For instance, HUMAN nouns will prototypically be the objects of communication verbs, but an example such as "McLeish has rung his own flat to collect messages" shows that the complements can undergo a particular assimilation process as to have a HUMAN interpretation. For our approach, this meant that there would also be noisy cases, marked contexts occasionally occurring with non-members of the class.

Besides, systematic polysemy has also to be taken into account: 'Book', for example, is SEMIOTIC, as well as EVENT. We addressed that semantic aspect by making our classes binary and assuming multi-classification. We consider class assignment based on a distributional approach to be a characterization of the semantics of words according to a particular corpus. Therefore, we can expect a certain deviation in regards to gold standard encoding that will much depend on the particular domain of the corpus. With all of these possible problems in mind, we have examined to what extent cue-based classification can overcome them.

On the light of this analysis, noun classification into lexical semantic classes appears to be a complex task. A number of linguistic cues must be identified as marks for each class. This task is language dependent as these marks might have different realizations for each of them. Besides, the members of a class are expected to appear also in non-marked contexts, just as the occurrence of non-members in the marked contexts is expected too. Although the classification is expected to have limited success, our challenge is to measure to what extent these results are effective for practical use; in particular, to what extent they reduce the amount of manual work in lexicon annotation. As in Bel et al. (2010), we have applied a confidence-based threshold yielding results that estimate this human work reduction in a 30-40% (section 5).

## 4. Cues for lexical semantic classes

In what follows, we report on the particular cues we have used for our experiments after a linguistic study of the relevance of possible marking contexts, as explained in section 3. According to our assumptions there, we have inspected the following range of linguistic phenomena for identifying efficient discriminating features:

*Predicate selectional restrictions*

Most verbs impose particular semantic restrictions to their subjects and objects: verbs like 'happen' and 'cause' are said to select different types of nouns as subjects, and these differences can be generalized under the lexical-semantic class concept. 'Happen' selects for EVENT nouns as subjects, whereas 'cause' selects for agentive entities, among which HUMAN.

Thus, HUMAN nouns in both English and Spanish can be identified as subjects of particular agentive verbs, and those that denote an intelligent act, such as *admire, talk, think,* etc. in English and in their corresponding expressions in Spanish.

Selectional restrictions also apply to complements other than the subject and object. In the case of LOCATION nouns, verbs imposing certain selectional restrictions impose also subcategorization frame constraints in the form of prepositional complements, although in this case the differences in the lexicalization of movement in both languages (Jackendoff, 1983) made the verb list different for the two languages considered. Thus for English, verbs such as 'come', 'go' and 'arrive' are used as cues with different prepositions, as the Spanish translations, 'venir',

'ir' and 'llegar'. For English we have also used motion verbs that do not require preposition: 'enter', 'leave', etc.

Selectional restrictions are also imposed by non-verbal predicative elements like adjectives that can restrict the nouns they combine with. While the strongest case is the case of collocations, there are classes of adjectives imposing constraints on the class of nouns they modify. For instance, Dixon (1982) identified 'human propensity' adjectives. We have also used co-occurrence with particular adjectives as cues for HUMAN (geographical provenance) and for LOCATION (adjectives such as 'far', 'remote', etc.).

*Grammatical Functions*

There are particular grammatical functions that also select for nouns that have particular semantic characteristics. While the class of the subject is largely determined by the selectional restrictions of the predicate, as we have just exemplified, we can say that Indirect Objects both in English and Spanish preferably select for HUMAN nouns, and to a certain extent, that by-Objects in passive constructions are also filled in by HUMAN nouns. HUMAN nouns are also related to the dative alternation phenomena in English. In addition, in Spanish, Direct Objects marked with the preposition 'a' are mostly HUMAN.

However, the role of grammatical functions as marks of particular classes depends on the class and, from that point of view only the HUMAN class seems to be correlated with particular grammatical functions. On the contrary, the class of LOCATION does not overtly correlate with any grammatical function, but it does with particular prepositions heading the prepositional phrases in which the noun occurs.

Adjuncts or modifiers of the nouns to be classified are also informative, as well as the occurrence of these nouns as modifiers in particular cases, if also combined with the occurrence of particular particles. Clear cases of modifiers that describe the semantic characteristics of the noun they modify are relative clauses headed with marked relative pronouns: 'who' and 'whom' ('quien' is the Spanish correlate), for example, clearly refer to a HUMAN antecedent, while 'where' (or the Spanish equivalent 'donde', which is more restricted to LOCATION than the English counterpart) are related to the class of LOCATION.

For English in particular, genitive complements ('my brother's book') are more often filled with nouns belonging to the HUMAN class. Also, possessive determiners tend to modify HUMAN nouns, as for instance in 'his colleagues'. However, and as expected, these cues can only be considered in correlation with other cues as they cannot be considered indicative just and only of the classes we are looking for. As suggested by usage grammar theories, the emergent classes are based on a number of marked correlations.

*Prepositions*

Prepositions, especially those said to be 'content' prepositions, that is, those that bear meaning, are also informative of the lexical semantic class of the noun filling the noun phrase they precede. 'During' and the corresponding Spanish 'durante' are key cues to identify members of the EVENT class. While prepositions such as 'at', 'within', 'across' or 'under' are good hints of LOCATION for English nouns. Some examples for Spanish are 'en' and 'según' ('in' and 'according to'), which are indicative of LOCATION and HUMAN nouns respectively.

Nouns themselves also bear complements and modifiers that are selected by the noun semantics. Depending on the language, they appear as noun-compounds or as PPs. Prepositions heading this complement PP are often not informative ('of', for instance).

*Suffixes*

Morphology is an important hint for several lexical semantic classes. Particular derivational affixes (Light, 1996) are good indicators of HUMAN nouns in both languages. In English for example, suffixes such as '-er', '-or', '-ist', etc. quite effectively identify HUMAN nouns, while in Spanish suffixes such as '-aco', '-ano', '-dor', etc. are good cues.

For LOCATION, suffixes in Spanish, are much more restricted to a class than in English. These cues are intended to register the locative nouns that contain discriminating suffixes such as '-dom', '-eria', '-place', etc., in English, and '-ería', '-al', '-dero', etc., in Spanish.

As we have seen, each of the classes was characterized by a number of different cues for each language that were manually identified following the guidelines mentioned before. Not all of them have the same distribution varying in sparseness (low frequency) and noise (also occurring with non-members of the class). In the annex (Tables 3-7), the reader can find the tables with the cues identified for each class and language, as well as their distribution in the corpora used in the experiments.

For our cue-based lexical classification tasks, the results of the cue *n*-pattern checking in all the occurrences of a word in a corpus were stored as features in an *n*-dimensional vector. As already explained, the sparse data problem is a heavy constraint for the actual effectiveness of the cues in learning and classification because very informative cues can be very infrequent in a corpus and therefore be ignored. In order to maximize the gathering of information, in addition to the mentioned cues, we have devised two additional strategies: firstly, for lexical co-occurrence, we have only used the 1000 first ranked words as extracted from a list of frequent words[1]. Secondly, we have collapsed cues of the same type in one single feature, for instance, different agentive verbs. Finally, because we expected to have noisy cases because of coercion, we included cues that were meant to be negative. These negative cues were an attempt to capture correlations with other marks that separate members of the non-members.

## 5. Experiments and evaluation

Our experiments have covered English and Spanish nouns for the following classes: EVENT, HUMAN, CONCRETE, SEMIOTIC, LOCATION and MATTER. In this paper, we mostly give details, for the sake of comparison, on the experiments for LOCATION, EVENT and HUMAN classes both for English and Spanish.

For our experiments, we used the *Corpus Tècnic de l'IULA* (Cabré et al., 2006). We used different English and Spanish corpora: for Spanish, a newspaper corpus of 21M tokens; for English, a corpus of 3.2M tokens consists of texts of different domains. We experimented with both Spanish and English in order to conduct a cross-linguistic analysis. We used a DT classifier in the Weka (Witten & Frank, 2005) implementation of pruned C4.5 DT (Quinlan, 1993). The DT performs a general to specific search in a feature space that selects the most informative attributes for a tree structure as the search proceeds. Here, the goal is to select for the minimum set of attributes that can efficiently partition the feature space into classes of observations and assemble them into a tree. In the experiment, we used a 10-fold cross-validation testing.

In regards to the gold-standard lists used for training and evaluation, we used already available, manually annotated lists of nouns extracted from the lexicon of a rule-based Machine Translation System (Alonso and Bocsák, 2005) for Spanish. For English, we created the gold standards using data from the SemEval 2007 workshop *Task 07: Coarse Grained English All-Words* (Navigli et al., 2007). The words used in this task were first automatically tagged with an automatic clustering method (Navigli, 2006) using senses based on the WordNet sense inventory and later manually validated by expert lexicographers. For our experiments, we extracted all of the words from this inventory that contained as their first sense a sense that corresponded to the lexical semantic classes, i.e. "people" in the case of the class HUMAN. The gold standards were not contrasted with the actual occurrences of the nouns in the corpora.

Gold-standards were in principle balanced with respect to class members and non-members, although the actual occurrences in the corpus determined the final lists. Thus, a baseline based on the majority class cannot be drawn from the gold-standards. A baseline based on the majority class in an actual dictionary will not be indicative as there will always be a majority of non-members.

[1] According to http://www.macmillandictionaries.com/

The following tables (1 and 2) show the results obtained in our experiments in terms of accuracy, False Positives (FP) and False Negatives (FN). Also, we show the best accuracy that can be obtained using a confidence threshold to select the elements that have been classified with the highest precision (around a 90%), and the revision manual work to be performed, i.e., the percentage of items in the gold-standard lexica that have been classified below the threshold and which would require human inspection.

| Class | Acc. (%) | FP (%) | FN (%) | Using confidence threshold | |
|---|---|---|---|---|---|
| | | | | Acc. (%) | To be revised (%) |
| HUM | 77.29 | 9.67 | 13.04 | 91.47 | 68.27 |
| LOC | 77.55 | 9.84 | 12.61 | 89.08 | 68.73 |
| EVENT | 80.90 | 6.53 | 12.56 | 92.85 | 66.33 |

Table 1: DT results for Spanish, including accuracy, percentage of false positives and false negatives and the assessment of entries to be revised.

| Class | Acc. (%) | FP (%) | FN (%) | Using confidence threshold | |
|---|---|---|---|---|---|
| | | | | Acc. (%) | To be revised (%) |
| HUM | 79.01 | 5.52 | 15.47 | 89.36 | 65.38 |
| LOC | 66.21 | 11.64 | 22.15 | 81.60 | 71.46 |
| EVENT | 73.05 | 8.38 | 18.56 | 83.33 | 71.26 |

Table 2: DT results for English, including accuracy, percentage of false positives and false negatives and the assessment of entries to be revised.

## 6. Discussion

Tables 1 and 2 present the results obtained from our experiments using linguistically selected cues. The overall results show that the selected cues can be informative in distinguishing the addressed lexical semantic noun classes. Hence, we confirmed our hypothesis that it is possible to exploit the correlation between syntactic, morphological and lexical co-occurrence to identify members of a lexical semantic class. From the results, we can observe differences that are related to the size of the corpus and what we call the grammaticalization degree of a particular class in a particular language, i.e. the availability of marks in the form of grammatical functions or morphological cues, which are more pervasive indicators than lexical co-occurrence, a sparser phenomenon.

In general, false positives, as expected, resulted from noisy cases, demonstrating that classes are marked by contexts that are identifiable as linguistic cues. In our experiments, noise can be related to the low-level tools used (Regular Expressions over PoS tagged corpora) and to some coercion contexts, as explained in section 3. Some examples of what we found to be noise are, for instance, the noun 'pancarta' ('banner'), which is found after the prepositional expression 'después de' ('after') perhaps referring to the temporal sequence of a demonstration headed by it in a clear coercion case. Another example is the noun 'cárcel' ('prison'), for which there are some occurrences of 'años de cárcel' ('years of prison') in the corpus. This would lead us to consider that 'prison' or 'banner' can be interpreted as an event or that the cue produces some undesired matching.

In turn, false negatives show that indeed the main problem is the lack of data. For example, there are 68 English HUMAN nouns (almost 13% of the total) that were not found in any of the contexts that were taken as cues. Experiments with English data achieved less accuracy than with Spanish data. However, note that though the English corpus is smaller (approx. 15%) than the Spanish corpus, the results confirm that the amount of information supplied by the corpus is indeed a success factor, though not as important as generally assumed.

For HUMAN and EVENT classes, the accuracy results are more similar in both languages. The size of the corpus seems to affect more the classification confidence than the general accuracy when comparing the threshold figures. However, it is noticeable that the size of the corpus is not the only factor that causes this lack of data.

The differences in the LOCATION class could signal the degree of grammaticalization. One of the clearest examples of this phenomenon within our experiment is for the HUMAN class with the case of Spanish that consistently marks human direct objects using the preposition 'a'.

The morphological cues, though applicable for both languages, have different results depending on the language and the class. For example, in English, derivational suffixes were strong marks for the HUMAN class, as many HUMAN nouns are nominalizations. However, this did not hold true in English for the LOCATION class. In this case, the derivational affixes are quite noisy, in comparison to Spanish. This could be attributed to the fact that in English, the LOCATION class relies heavily on compounding such as "rice field' and 'rose garden' in English but 'arroz*al*' and 'rosal*eda*' in Spanish.

## 7. Conclusions

The results of our experiments support our hypothesis about the use of cues to classify nouns in lexical semantic classes, despite the complexities that the task contained. The hypothesis that cue correlations, more than particular, exclusive cues, provided a strong predicative power seems also to have been confirmed because none of the used cues prove to be exclusive of the class. We also have shown our methodology to be language-independent, though the cues themselves are not. While the accuracy results are not conclusive, it can be mostly attributed to the sparse data problem, according to the higher rate of false negatives than false positives.

Overall, we can also conclude that with respect to a manually created dictionary, using the method just reported we have done between 30 to 40% of the work automatically. Further work includes extending this framework to include more classes and languages.

## 8. Acknowledgements

This work was funded by the EU 7FP project 248064 PANACEA and the UPF-IULA PhD grant program.

## 10. Annex

The following tables give distributional information about the actual cues used in our experiments. The X represents where the class members should be found.

|   | Cues | | Relative Frequency | |
|---|---|---|---|---|
|   |   |   | Location | Non Location |
| 1 | Suffix | -eria, -ion, -ity, -dom, -ory, -topy, -ium, -ile, -polis, -way, -ment, -sphere | 0.01929 | 0.02463 |
| 2 | X_where | | 0.00623 | 0.00310 |
| 3 | at_X | | 0.02491 | 0.00405 |
| 4 | in_X | | 0.09748 | 0.08755 |
| 5 | outside_X | | 0.00073 | 0.0 |
| 6 | across_X | | 0.00093 | 0.00015 |
| 7 | from_X | | 0.01724 | 0.00775 |
| 8 | to_X | | 0.03313 | 0.02424 |
| 9 | along_X | | 0.00091 | $5 \cdot 10^{-5}$ |
| 10 | inside_X | | $1 \times 10^{-5}$ | $2 \times 10^{-5}$ |
| 11 | through_X | | 0.00157 | 0.001395 |
| 12 | toward_X | | 0.00064 | 0.00049 |
| 13 | within_X | | 0.00602 | 0.00049 |
| 14 | leave (V)_X | | 0.00064 | 0.00020 |
| 15 | come(V)_PP_X | | 0.00214 | 0.00175 |
| 16 | Modifiers with suffix: Spanish_X | | 0.00554 | 0.00170 |
| 17 | Adjectives of dimension: distant_X | | 0.00570 | 0.00222 |
| 19 | X_think 'Agentive verbs' | | 0.00162 | 0.00219 |
| 20 | use_X 'non-loc objects' | | 0.01022 | 0.01266 |
| 21 | with_X | | 0.013379 | 0.01662 |

Table 3: Cues for LOCATION class in English

|   | Cues | | Relative Frequency | |
|---|---|---|---|---|
|   |   |   | Location | Non Location |
| 1 | Suffix | -dromo, -puerto, -dor, -dero, -ería, -orio, -al, -teca, -polis, -edo, | 0.07280 | 0.01782 |
| 2 | X_donde (X_'where) | | 0.00484 | 0.00028 |
| 3 | en_X ('in X') | | 0.17222 | 0.08249 |
| 4 | hacia_X ('toward_x') | | 0.00373 | 0.00083 |
| 5 | hasta_X ('until_X') | | 0.00269 | 0.00111 |
| 6 | desde_X ('from_X') | | 0.00636 | 0.00178 |
| 7 | entre_X ('between_X) | | 0.00411 | 0.00410 |
| 8 | entrar_X ('to enter X') | | 0.01865 | 0.00407 |
| 9 | Modifiers with suffix: X_español ('Spanish_X') | | 0.02074 | 0.01041 |
| 10 | Adjectives of dimension: X_cercano ('close_X') | | $2 \cdot 10^{-5}$ | 0.0 |
| 11 | X_Modal Adj. ('X_possible') | | 0.00031 | 0.00090 |
| 12 | Modal Adj._X ('posible_X') | | 0.00102 | 0.00181 |

Table 4: Cues for LOCATION class in Spanish

|   | Cues | | Relative Frequency | |
|---|---|---|---|---|
|   |   |   | Human | Non Human |
| 1 | Suffix | -er, –or, –man, –men, –mate, –ist, –arian, –naut, –yst, –ster, –ess, –ist, –ant, -ian | 0.53867 | 0.07553 |
| 2 | Prefix | pro-, grand- | 0.06680 | 0.04336 |
| 3 | X_decide (V) | | 0.00943 | 0.00510 |
| 5 | by_X | | 0.02340 | 0.01058 |
| 6 | V_N_to_X (Indirect Objects) | | 0.00786 | 0.00502 |
| 7 | V_X_N (Direct Object) | | 0.00749 | 0.00728 |
| 8 | genitive X_N | | 0.01868 | 0.00115 |
| 9 | X, who | | 0.00934 | 0.00095 |
| 10 | group of X | | 0.00129 | 0.00061 |
| 11 | jealous (Adj.) X | | 0.00952 | 0.00206 |
| 12 | to begin_X | | 0.00064 | 0.00202 |
| 13 | during_X | | 0.00083 | 0.00732 |
| 15 | much_X | | 0.00027 | 0.00061 |

Table 5: Cues for HUMAN class in English

|   | Cues | | Relative Frequency | |
|---|---|---|---|---|
|   |   |   | Human | Non Human |
| 1 | Suffix | -aco, -ano, -ario, -cida, -crata, -cultor, -dor, -eco, -ego, -eño, -ero, -és, -ista, -nte, -ólogo, -pata, -quía, -triz, -uta | 0.36652 | 0.02612 |
| 2 | Prefix | re-, sobre-, sub- | 0.00099 | 0.00067 |
| 3 | X_pretender ('intend') | | 0.00070 | 0.00035 |
| 4 | X_tener ('X_to have') | | 0.00071 | 0.00050 |
| 5 | por_X ('by_X') | | 0.01971 | 0.01762 |
| 6 | IO_clitic V a_X ('IO_clitic V to_X') | | 0.00225 | 0.00093 |
| 7 | V_a_X ('V to X') | | 0.04750 | 0.02134 |
| 8 | possesive_X | | 0.05912 | 0.04667 |
| 9 | según_X ('according') | | 0.00404 | 0.00134 |
| 10 | sobre_X ('about X') | | $3 \cdot 10^{-5}$ | $9 \cdot 10^{-5}$ |
| 11 | Noun_X | | 0.01522 | 0.01096 |
| 12 | X se V | | 0.01320 | 0.00986 |
| 13 | enchufar X ('to plug X') | | 0.00032 | 0.00120 |
| 14 | desde_X ('from_X') | | 0.00076 | 0.00304 |
| 15 | hacia_x ('towards_X') | | 0.00068 | 0.00096 |
| 16 | durante_X ('during_X') | | $2 \cdot 10^{-5}$ | 0.00098 |
| 17 | V_X | | 0.04537 | 0.10826 |

Table 6: Cues for HUMAN in Spanish